\documentclass{article}

 \PassOptionsToPackage{numbers, compress}{natbib}
 \usepackage[final]{nips_2017}
\usepackage{graphicx}
\usepackage[utf8]{inputenc} 
\usepackage[T1]{fontenc}    
\usepackage{hyperref}       
\usepackage{url}            
\usepackage{booktabs}       
\usepackage{amsfonts}       
\usepackage{amsmath}       
\usepackage{nicefrac}       
\usepackage{microtype}      
\usepackage{multirow}      

\title{Delineation of Skin Strata in Reflectance Confocal Microscopy Images using Recurrent Convolutional Networks with Toeplitz Attention}

\author{
  Alican Bozkurt, Jaume Coll-Font, Dana H. Brooks, Jennifer G. Dy \\
  Northeastern University\\
  Boston, MA 02115 \\
  \And
  Kivanc Kose, Milind Rajadhyaksha \\
  Memorial Sloan Kettering Cancer Center\\
  NYC, NY 10022 \\
  \And
  Christi Alessi-Fox \\
  Caliber I.D. Inc.\\
  Rochester, NY 10022 \\
}

\begin{document}

\maketitle

\begin{abstract}
Reflectance confocal microscopy (RCM) is an effective, non-invasive pre-screening tool for skin cancer diagnosis, but it requires extensive training and experience to assess accurately. There are few quantitative tools available to standardize image acquisition and analysis, and the ones that are available are not interpretable. In this study, we use a recurrent neural network with attention on convolutional network features. We apply it to delineate skin strata in vertically-oriented stacks of transverse RCM image slices in an interpretable manner. We introduce a new attention mechanism called Toeplitz attention, which constrains the attention map to have a Toeplitz structure. Testing our model on an expert labeled dataset of 504 RCM stacks, we achieve $88.17\%$ image-wise classification accuracy, which is the current state-of-art.
\end{abstract}

\section{Introduction}
Reflectance confocal microscopy (RCM) is a non-invasive optical imaging technology that enables users to examine mosaics of small field of view (FOV) images of $1.5\,\mu m$ thick layers ("optical sections") of skin at $0.5\,\mu m$ lateral resolution. Imaging can go as deep as $200\,\mu m$, which is sufficient for diagnosing important skin conditions,  typically covering the epidermis and papillary dermis. Recent studies have demonstrated that RCM imaging is highly sensitive ($90-100\,\%$) and specific ($70-90\,\%$) for detecting skin cancers~\cite{Raj16} by expert visual inspection. Moreover, the combination of RCM and dermoscopy has been shown to reduce the rate of biopsy of benign lesions per detected malignancy by ${\sim}\,2\times$, leading to better patient care~\cite{Borsari16,Pell16}.

\begin{figure}[ht]
    \centering
    \includegraphics[width=0.9\linewidth]{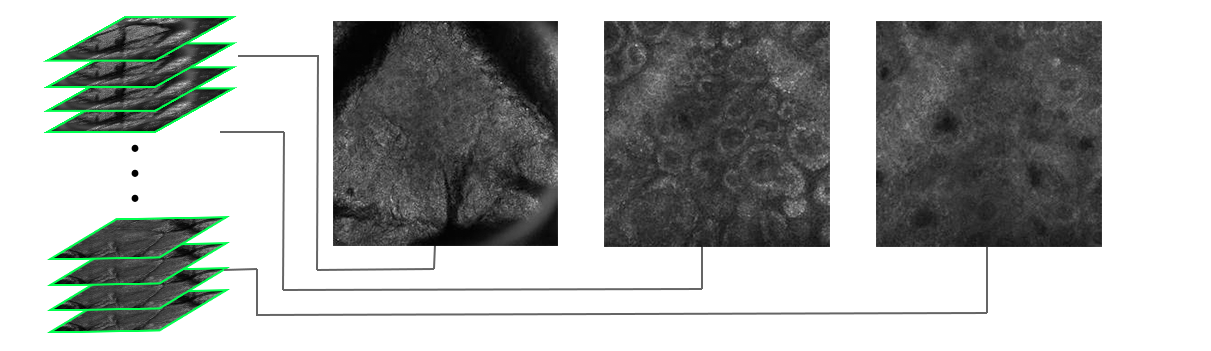} 
    \caption{\textbf{An example stack} with epidermis(left), DEJ (middle), and epidermis (right) images that show how appearance changes as depth increases}
    \label{fig:stack-mosaic}
\end{figure}

To identify the correct depths for gathering mosaics, clinicians take a sequence of RCM images at $1 - 5\,\mu m$ separation in depth, from the epidermis to the dermis. This set of images is referred to as an RCM \textit{stack} and each RCM image in the stack is called an RCM \textit{slice}. After acquiring the stack, the clinician classifies each image as belonging to an epidermis, dermal-epidermal junction (DEJ), or dermis layeer. Figure~\ref{fig:stack-mosaic} shows typical RCM images from each layer. The clinician then uses the stack as a reference to choose depths at which to collect larger FOV high-resolution images mosaics for subsequent diagnostic analysis.

Our work is concerned with the first part of the process. Given a stack of RCM images, we try to distinguish boundaries between the epidermis, dermal-epidermal junction, and dermis. RCM stack data has a strong sequential structure, which recurrent neural networks can naturally exploit. Human skin maintains a strict ordering of different strata; there is a smooth transition between the contiguous layers (e.g. epidermis$\rightarrow$dermis$\rightarrow$epidermis transitions do not happen) and unidirectional (dermis$\rightarrow$DEJ or DEJ$\rightarrow$epidermis transitions are not possible). These constraints help experienced clinicians quickly classify the stacks. This observation was also made in \cite{bozkurt2017delineation}, where authors used a recurrent convolutional network (RCN) to emulate the clinicians' behavior. They reported high accuracy and a consistent classification by using a model that looks at all of the images in a stack. They also hinted at the possibility of performing equally well considering only a small part of the stack at a time. They tested this possibility by performing the same task while only looking at three slices at a time, and reported  only minimal loss in terms of performance.

In this work, we apply two attention-based models to extend \citep{bozkurt2017delineation}. Our first model is similar to their full-sequence RCN model, in the sense that both approaches use information from all images in a stack. Even though their model has very powerful representation ability (and was shown to perform very well for this dataset), it is not interpretable: we do not know how much each image is contributing to a decision. To improve the interpretability of their model, we we utilize a soft (global) attention mechanism \cite{bahdanau2014neural}, which also uses information from all images in a stack, but applies a soft mask over representation of images in the network. Visualizing the mask gives us clues about which images the network pays attention to which making a decision.    

We also address redundancy in  \cite{bozkurt2017delineation}. With their partial-sequence RCN model, the authors showed that it is possible to perform comparably to full-sequence RCN model, just by looking at 3 images per decision. However, choice of number of images to use is not well justified. Finding the optimal location and number of components of an input to look at is a well known problem \citep{mnih2014recurrent} in computer vision, and solutions like hard-attention require sample approximation algorithms like REINFORCE \cite{williams1992simple}, due to the discrete nature of the problem. To find a compromise between that and soft attention, we restrict the support of the attention vector and make it sequence-independent, which results in a Toeplitz structured attention map. In the following section, we will explain the attention mechanisms used in this work in more detail. 
\section{Global Attention}
Global attention \cite{bahdanau2014neural} has been proposed as a way to align source and target segments in neural machine translation in a differentiable manner. Since then, it has been used in many computer vision and NLP related tasks. In this mechanism, an attention vector with the same length as the sequence is calculated with a neural work from encoding of the whole sequence, then a context vector is calculated as weighted sum of encodings according to the attention vector. 

\section{Toeplitz Attention}
The name Toeplitz Attention comes from the idea that the attention map created by this method has Toeplitz structure. This method is a compromise between global attention and soft attention. Support of the attention weights is more compact than global attention, but it is still differentiable.

This mechanism can be seen as a special case of local attention with monotonic alignment introduced in \citet{luong2015effective}, where the context vector was calculated as a weighted average over sets of $\mathbf{h}_t$ within the window $t \in [p_t - D, p_t+D]$ ($D$ is chosen, in both \cite{luong2015effective} and our work, empirically). Aligned position $p_t$ can be calculated with an MLP (predictive alignment) or set as $p_t=t$ (monotonic alignment). Monotonic alignment is suitable when source and target sequences are aligned, such as our case. Now $\mathbf{a}_t$ has a shorter support of $2D+1$, compared to the full input sequence length in the global attention case. It is calculated in a similar fashion to global attention in \cite{luong2015effective}. In our case $\mathbf{a}_t$ is time (here, depth) independent, i.e. $\mathbf{a}_t=\mathbf{a}$ where $\mathbf{a}$ is a learnable kernel with all non-negative entries that sum to one (convex combiner). The attention map $A$, which is a concatenation of $\mathbf{a}_t$ for each slice, $A=[{a}_1;\ldots;{a}_T]$ therefore will have a Toeplitz structure. This structure lends itself to an efficient implementation using convolution.      

\begin{figure}[ht]
    \centering
    \includegraphics[width=0.4\linewidth]{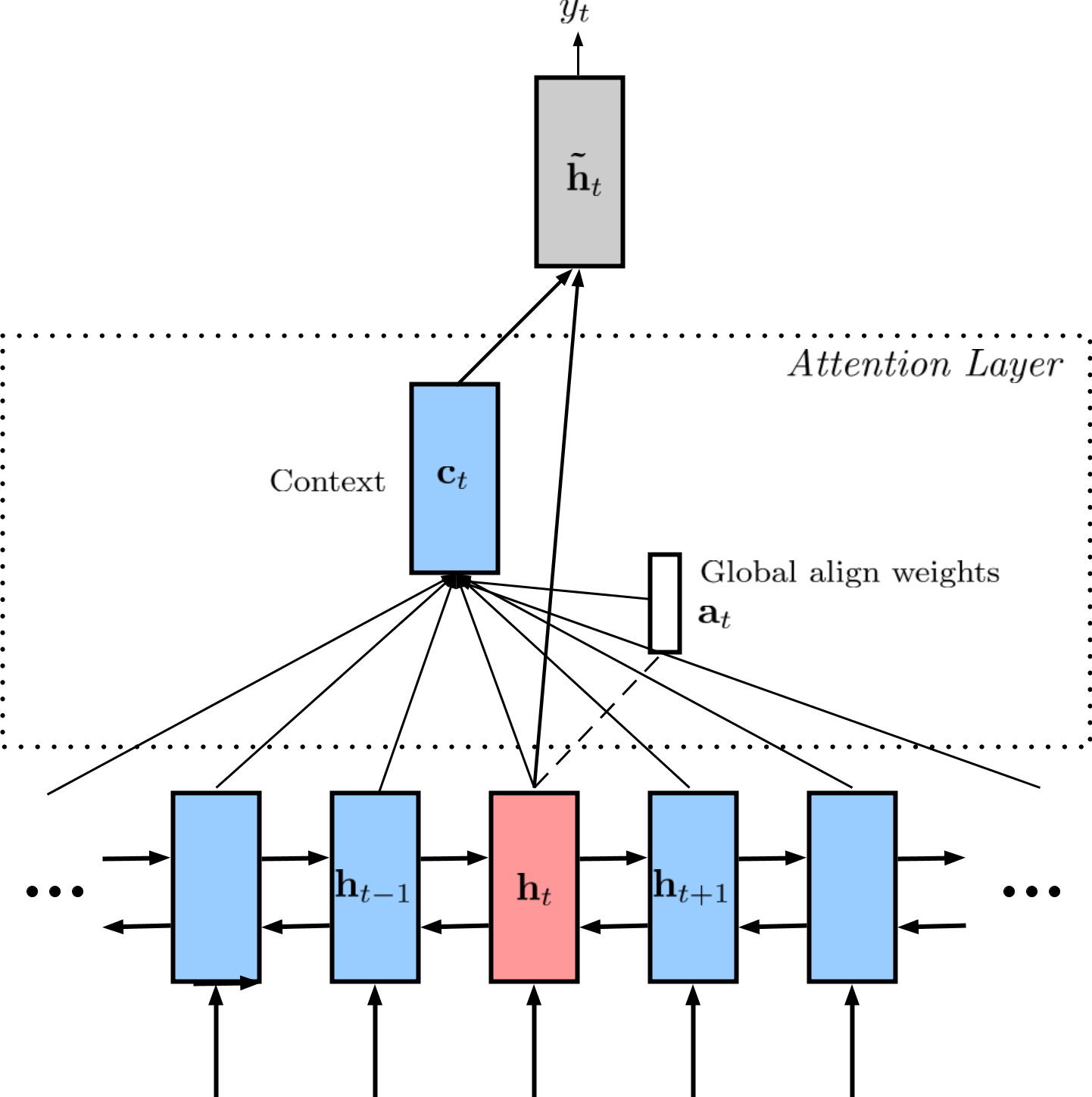} 
    \includegraphics[width=0.4\linewidth]{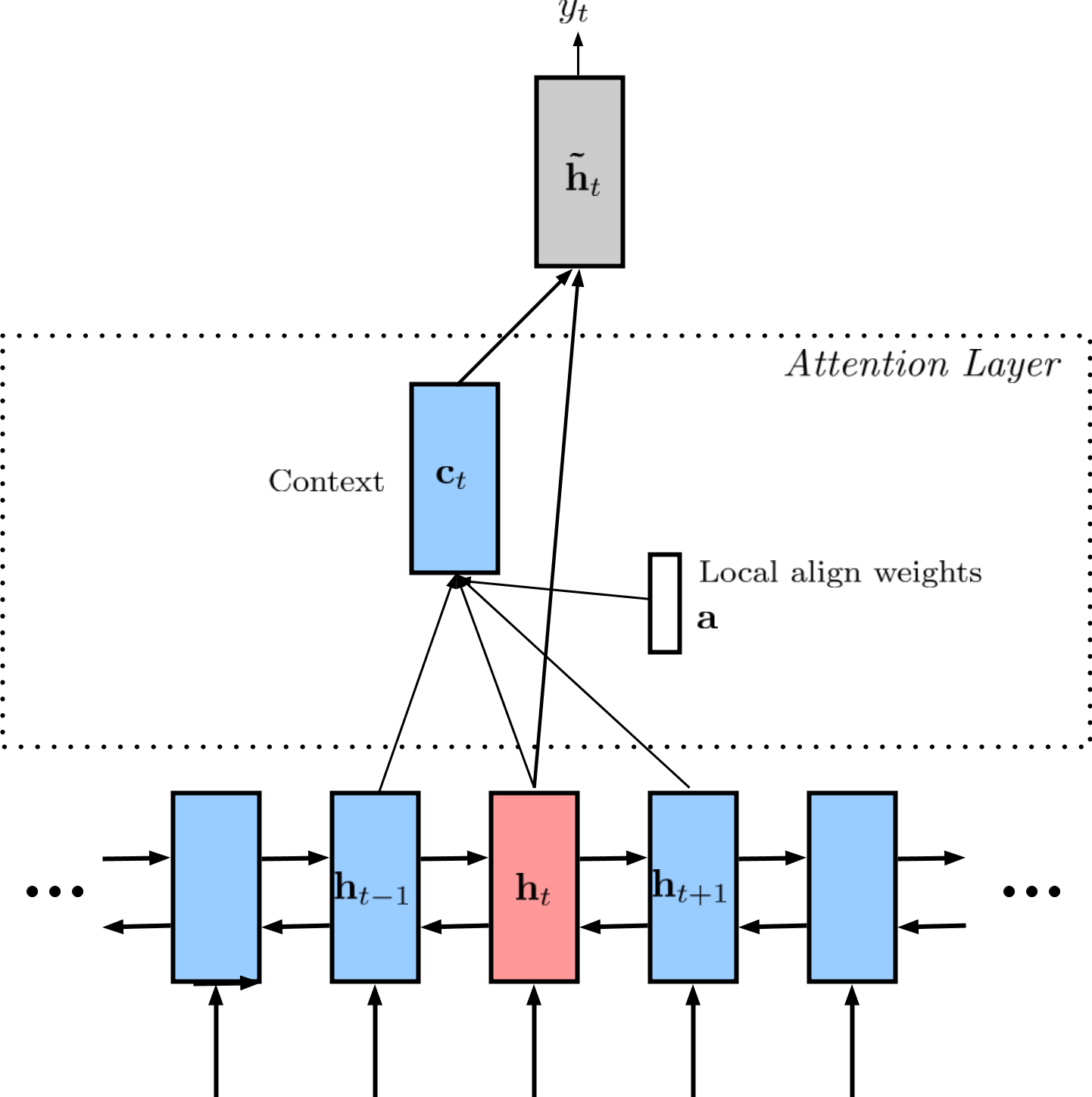}
    \caption{\textbf{Attention mechanisms.} (left) Global attention model. (right) Toeplitz attention model. Note that this figure is intended to explain only the attention layers, the encoder and decoder structures are not presented in detail due to space constraints.}
    \label{fig:att}
\end{figure}

\section{Model Definition}
In the neural machine translation literature, the attention layer is typically applied between an encoder and decoder \cite{cho2014learning, sutskever2014sequence}. We  replicate this structure in our work as well. We  use bidirectional gated recurrent units (GRU)\cite{GRU} appended to Inception v3 networks \cite{inceptionV3} to create a recurrent encoder network. This network will produce encoding for every image in a stack. The full sequence RCN model used in \cite{bozkurt2017delineation} can be formed by attaching a fully connected layer to end of this encoder. From this lens, it can be seen as a special case of attention augmented networks. Indeed, we can recover the full-sequence RCN as a special case of Toeplitz attention where $D=0$, so the attention map becomes an identity matrix.

We use different decoder networks for each attention mechanism. For global attention, we use a GRU followed by a fully connected layer. For Toeplitz attention, we use a just a fully connected layer. In both cases, we augment the attended input to the decoder by the decoder's output at a previous timestep (again, time corresponds to slice depth here), to efficiently exploit the structure nature of the data. 

\section{Results}
We tested our methods on the dataset in \citet{bozkurt2017delineation}. To compare our methods with current state-of-art, we present accuracy, sensitivity, specificity, and number of errors that imply impossible transitions. Perhaps the most interesting parameter of our model is $D$, the support of the local attention vector. We experimented with several values to provide insight about the model. For $D=0$, we can compare our network with Toeplitz attention to full-sequence RCN to assess the effect of input-feeding.The $D=1$ case can be compared with partial-sequence RCN, however the main difference between models is that partial-sequence RCN's attention acts on the input (output of Inception v3 in their case) and operates in a complex manner that cannot be expressed with a weighted sum (due to recurrent units). 

Interestingly, looking at 3 images per decision indeed gives the highest accuracy in our experiments (Table~\ref{tab:acc}). Even though the global attention model does not perform as accurately as the Toeplitz attention models, it is the only model that performs perfectly consistently in terms of producing no anatomically impossible transitions, Table~\ref{tab:errors}). 

Looking at attention maps, we see the power of our model's capturing the sequential relationships within the stacks. The global attention model has a block-diagonal structure,and block edges align with transition boundaries. In the case of Toeplitz attention ($D=7$), even though the support of the attention vector is 15 slices, it learns to fit a sharper Gaussian-like curve around image of interest without any sparsity regularization. Note that the brightest band is not exactly on the main diagonal, so the network leverages the freedom to change the alignment. This freedom is restricted as $D$ decreases and the attention map for $D=0$ is an identity matrix, as expected.

\begin{table}[htbp]
\caption{Accuracy, sensitivity, and specificity for each class. Values for experiments not conducted in this work are taken from \cite{bozkurt2017delineation}}
  \centering
    \resizebox{0.9\textwidth}{!}{
    \begin{tabular}{|c|c|ccc|ccc|}
    \hline
    \multirow{2}[0]{*}{\textbf{Method}} & \multirow{2}[0]{*}{\textbf{Accuracy (\%)}} & \multicolumn{3}{c|}{\textbf{Sensitivity}} & \multicolumn{3}{c|}{\textbf{Specificity}} \\
    \cline{3-8}
          &       & \textbf{Epidermis} & \textbf{DEJ} & \textbf{Dermis} & \textbf{Epidermis} & \textbf{DEJ} & \textbf{Dermis} \\
    \hline
    Toeplitz Attention (D=1) & \textbf{88.18} & 93.76 & 83.88 & \textbf{84.34} & \textbf{95.98} & 90.48 & 95.75 \\
    Toeplitz Attention (D=0) & 88.04 & 93.88 & 83.55 & 84.12 & 95.71 & 90.82 & 95.45 \\
    Full seq. RCN \cite{bozkurt2017delineation}& 87.97 & 93.95 & 83.22 & 84.16 & 95.82 & 90.54 & 95.51 \\
    Toeplitz Attention (D=7) & 87.69 & 93.84 & 83.21 & 83.27 & 95.29 & 90.07 & 95.96 \\
    Par. seq. RCN \cite{bozkurt2017delineation}& 87.52 & 94.14 & 82.54 & 83.33 & 94.78 & \textbf{90.83} & 95.44 \\
    Global Attention & 86.47 & \textbf{95.54} & 81.38 & 78.89 & 93.94 & 89.19 & 96.20 \\
    Inception-V3 \cite{bozkurt2017delineation}& 84.87 & 88.83 & \textbf{84.66} & 78.18 & 95.84 & 85.73 & \textbf{96.23} \\
    \citet{Sam2016} & 84.48 & 88.87 & 80.93 & 81.85 & 93.81 & 87.81 & 94.78 \\
    \citet{Kaur16a} & 64.33 & 73.99 & 51.14 & 68.27 & 86.22 & 74.85 & 84.89 \\
    \hline
    \end{tabular}}%
  \label{tab:acc}%
\end{table}%

\begin{table}[htbp]
\caption{Anatomically impossible errors. Values for experiments not conducted in this work are taken from \cite{bozkurt2017delineation}}
  \centering
    \resizebox{0.6\textwidth}{!}{
    \begin{tabular}{|c|cccc|c|}
    \hline
    \multirow{3}[0]{*}{\textbf{Method}} & \multicolumn{4}{c|}{\textbf{Error Types}} & \multirow{3}[0]{*}{\textbf{Total}} \\
    \cline{2-5}
          & \multicolumn{1}{c}{Epidermis} & DEJ$\rightarrow$& Dermis$\rightarrow$& Dermis &  \\
          & $\rightarrow$Dermis & Epidermis & Epidermis & $\rightarrow$DEJ &  \\
          \hline
    Global Attention & \multicolumn{1}{c}{0} & 0     & 0     & 0     & 0 \\
    Toeplitz Attention (D=7) & \multicolumn{1}{c}{0} & 2     & 0     & 0     & 2 \\
    Toeplitz Attention (D=1) & \multicolumn{1}{c}{0} & 4     & 0     & 1     & 5 \\
    Full seq. RCN \cite{bozkurt2017delineation}& \multicolumn{1}{c}{0} & 4     & 0     & 3     & 7 \\
    Toeplitz Attention (D=0) & \multicolumn{1}{c}{0} & 2     & 2     & 4     & 8 \\
    Par. seq. RCN \cite{bozkurt2017delineation} & \multicolumn{1}{c}{3} & 10    & 5     & 5     & 23 \\
    Inception-V3\cite{bozkurt2017delineation} & \multicolumn{1}{c}{3} & 25    & 8     & 32    & 68 \\
    \citet{Sam2016} & \multicolumn{1}{c}{14} & 59    & 11    & 56    & 140 \\
    \citet{Kaur16a} & \multicolumn{1}{c}{32} & 255   & 16    & 99    & 402 \\
    \hline
    \end{tabular}}%
  \label{tab:errors}%
\end{table}%

\begin{figure}[h!]
    \centering
    \includegraphics[width=0.24\linewidth]{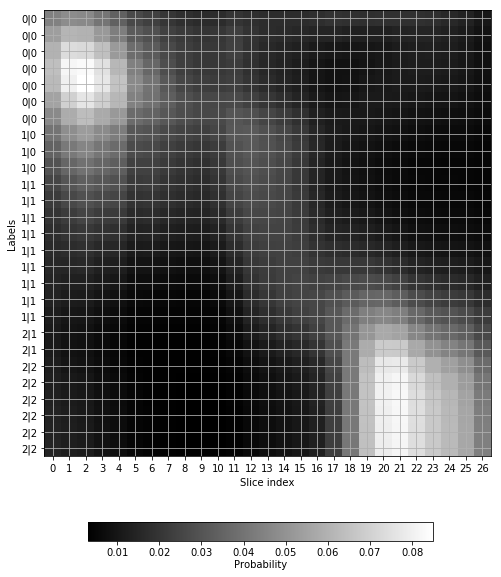} 
    \includegraphics[width=0.24\linewidth]{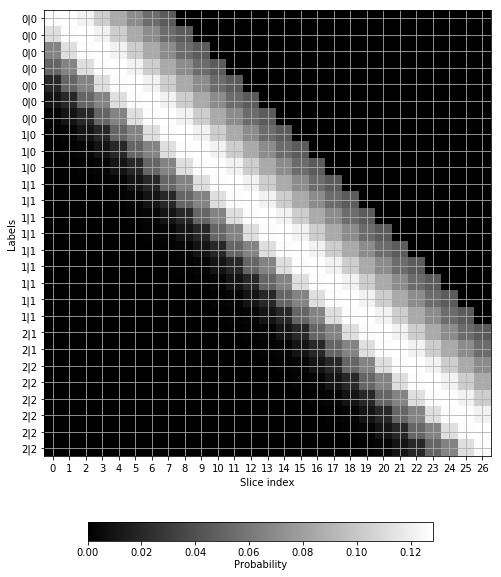}
    \includegraphics[width=0.24\linewidth]{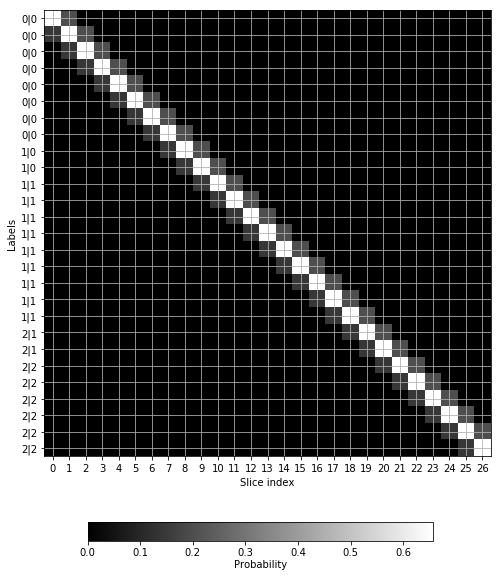}
    \includegraphics[width=0.24\linewidth]{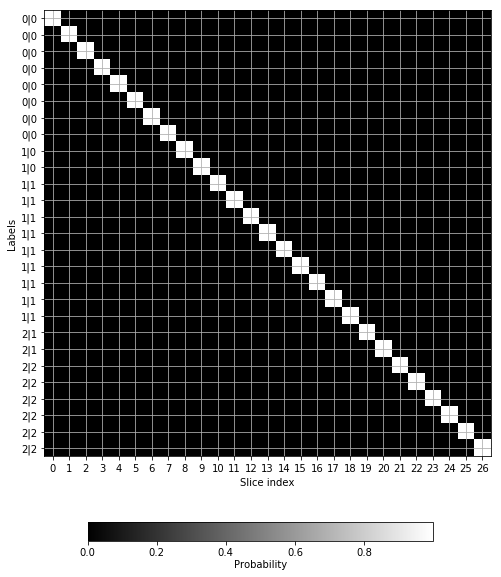}
    \caption{\textbf{Attention maps for different models} From left to right: global attention, Toeplitz attention ($D=7$), Toeplitz attention ($D=1$), Toeplitz attention ($D=0$)}
    \label{fig:attentionmaps}
\end{figure}

\section{Conclusion}
In this work, we incorporated attention mechanism to improve the interpretability of the recurrent convolutional networks of \citep{bozkurt2017delineation}. We experimented with two different mechanisms: first we tried global attention, where the network attends the whole stack with different weights. Second, we tried a Toeplitz attention mechanism, where we forced the attention map into a Toeplitz structure by making attention weights have smaller support and be depth-independent. Comparing Toeplitz attention with different $D$ parameters, we observed that indeed looking at 3 images per decision ($D=1$) gives the highest image-wise classification. Our model with global attention, in turn, behaves most consistently by reporting no anatomically impossible transitions. Comparing the $D=0$ case with the full-sequence RCN model, we also conclude that input-feeding in the decoder helps improve the image-wise classification accuracy, but does not help consistency.     

\section{Acknowledgements}
This project was supported by NIH grant R01CA199673 from NCI. This project was also supported in part by MSKCC's Cancer Center core support NIH grant P30CA008748 from NCI.

\bibliography{refs}
\bibliographystyle{plainnat}

\end{document}